\documentclass{article}

\PassOptionsToPackage{numbers, compress}{natbib}

\usepackage{amsmath} 
\usepackage{booktabs}
\usepackage{makecell}
\usepackage[preprint]{neurips_2026}
\usepackage{multirow}
\usepackage{graphicx}

\usepackage[utf8]{inputenc} 
\usepackage[T1]{fontenc}    
\usepackage{hyperref}       
\usepackage{url}            
\usepackage{booktabs}       
\usepackage{amsfonts}       
\usepackage{nicefrac}       
\usepackage{microtype}      
\usepackage{xcolor}         

\usepackage{cleveref}  %

\crefname{section}{Sec.}{Secs.}
\Crefname{section}{Section}{Sections} 

\crefname{table}{Tab.}{Tabs.}
\Crefname{table}{Table}{Tables} 

\crefname{figure}{Fig.}{Figs.}
\Crefname{figure}{Figure}{Figures}
\title{CausalCine: Real-Time Autoregressive Generation for Multi-Shot Video Narratives}

\author{%
  \textbf{Yihao Meng}$^{\heartsuit,1,2}$ \quad
  \textbf{Zichen Liu}$^{\heartsuit,1,2}$ \quad
  \textbf{Hao Ouyang}$^2$ \quad
  \textbf{Qiuyu Wang}$^2$ \quad
  \textbf{Ka Leong Cheng}$^2$ \\[0.15cm]
  \textbf{Yue Yu}$^{1,2}$ \quad
  \textbf{Hanlin Wang}$^{1,2}$ \quad
  \textbf{Haobo Li}$^{3,2}$ \quad
  \textbf{Jiapeng Zhu}$^2$ \quad
  \textbf{Yanhong Zeng}$^2$ \\[0.15cm]
  \textbf{Xing Zhu}$^2$ \quad
  \textbf{Yujun Shen}$^2$ \quad
  \textbf{Qifeng Chen}$^1$ \quad
  \textbf{Huamin Qu}$^1$ \\[0.4cm]
  \normalfont 
  $^1$HKUST \qquad
  $^2$Ant Group \qquad
  $^3$SJTU \\[0.2cm]
  \normalfont \small 
  $^\heartsuit$ Equal contribution
}
\begin{document}

\maketitle

\begin{abstract}
Autoregressive video generation aims at real-time, open-ended synthesis. Yet, cinematic storytelling is not merely the endless extension of a single scene; it requires progressing through evolving events, viewpoint shifts, and discrete shot boundaries. Existing autoregressive models often struggle in this setting. Trained primarily for short-horizon continuation, they treat long sequences as extended single shots, inevitably suffering from motion stagnation and semantic drift during long rollouts. To bridge this gap, we introduce \textbf{CausalCine}, an interactive autoregressive framework that transforms multi-shot video generation into an online directing process. CausalCine generates causally across shot changes, accepts dynamic prompts on the fly, and reuses context without regenerating previous shots. To achieve this, we first train a causal base model on native multi-shot sequences to learn complex shot transitions prior to acceleration. We then propose Content-Aware Memory Routing (CAMR), which dynamically retrieves historical KV entries according to attention-based relevance scores rather than temporal proximity, preserving cross-shot coherence under bounded active memory. Finally, we distill the causal base model into a few-step generator for real-time interactive generation. Extensive experiments demonstrate that CausalCine significantly outperforms autoregressive baselines and approaches the capability of bidirectional models while unlocking the streaming interactivity of causal generation. Demo available at \textcolor{magenta}{\href{https://yihao-meng.github.io/CausalCine/}{Project Page}}.

\end{abstract}

\section{Introduction}
\label{sec:intro}
Recent diffusion video models achieve impressive visual fidelity
\cite{seedance2026seedance,wan2025wan,hacohen2026ltx}, but their bidirectional
attention makes long, interactive generation expensive. Autoregressive generation
with KV caching offers a natural alternative for streaming video synthesis
\cite{huang2025liveavatar,bruce2024genie}, yet existing causal video models are
still largely trained and evaluated as short-horizon continuation systems
\cite{yin2025causvid,huang2025sf,zhu2026cf,yang2025longlive,liu2025rolling}.
When rolled out beyond a single local motion pattern, they often stagnate, loop,
or drift semantically \cite{cai2026mmm}. Cinematic long-form video, however, is
not merely an extended single shot. It requires evolving events, viewpoint
changes, discrete shot boundaries, and persistent story context.
\begin{figure}[t]
    \centering
    \includegraphics[width=\linewidth]{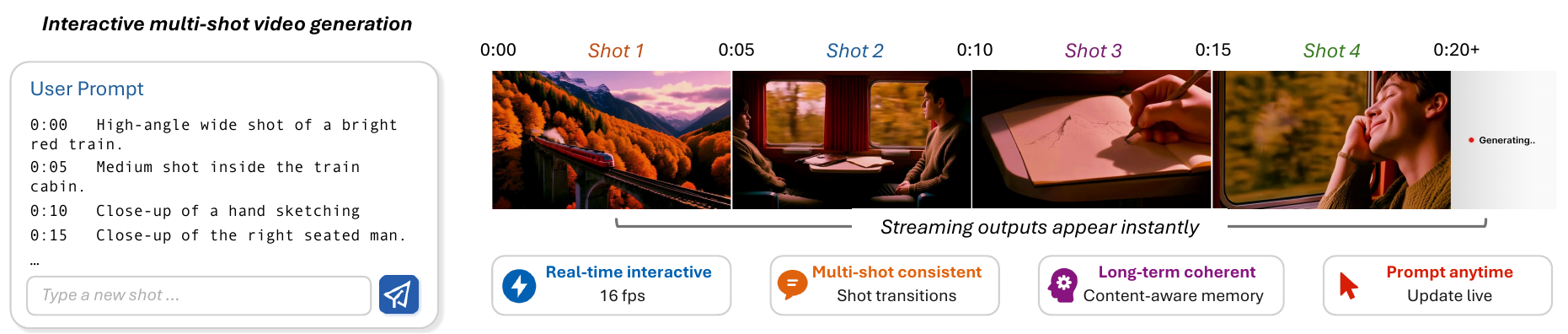}
       \caption{Real-time interactive multi-shot generation. CausalCine streams video chunks causally, accepts new shot-level prompts during rollout, and reuses content-aware KV memory so later shots can recall earlier content without regenerating previous video.}
    \label{fig:teaser}
    \vspace{-10pt}
    
\end{figure}

In this work, we study interactive autoregressive multi-shot video generation, where a model generates videos causally across shot changes, accepts new prompts during generation, and reuses relevant long-range context without regenerating previous shots. This setting exposes a key limitation of short-clip autoregressive training: the model must not only continue local motion, but also introduce new content at requested shot changes, follow newly appended prompts, and determine which information from earlier shots should remain accessible.

Our first observation is that long-form causal behavior should be learned before acceleration.
Instead of directly distilling a bidirectional diffusion model into a fast autoregressive generator \cite{yin2025causvid,huang2025sf}, we first train a full-step causal multi-shot base model on native long-form sequences with teacher forcing. The model observes shot boundaries, changing prompts, and long-range entity reappearance under the same causal dependency structure used at inference with KV caching. We find that high-quality native multi-shot data substantially reduces the usual teacher-forcing rollout gap, yielding a causal base model that can perform stable long rollouts and synthesize new content across shot transitions.

Autoregressive multi-shot generation also poses a greater challenge for KV memory. In single-scene continuation, fixed anchors or sliding windows can preserve local appearance and motion continuity~\cite{yang2025longlive,huang2025liveavatar,liu2025rolling}. However, when
generation must introduce new content, viewpoints, or environments, useful context is no longer determined by temporal proximity or fixed frame positions. The model may need to recall a character from the distant past, ignore the immediately preceding scene, or combine semantic cues from multiple earlier shots. We therefore introduce Content-Aware Memory Routing (CAMR), which selects historical KV entries by content relevance rather than fixed temporal position. CAMR retrieves useful long-range context and maintains a streamlined memory representation, improving cross-shot coherence without sacrificing causal generation.

Finally, we distill the causal multi-shot base model into a few-step generator for real-time interactive synthesis. Because causality and multi-shot structure have already been learned by the full-step model, Distribution Matching Distillation (DMD)
\cite{yin2024dmd,yin2024dmd2} can focus on trajectory compression while preserving visual quality and cross-shot consistency. The resulting model generates videos chunk by chunk with KV caching, supports prompt updates during generation, and continues a sequence without recomputing previous shots.

The resulting system enables real-time online directing for long-form video generation. Rather than rendering a complete video offline, CausalCine streams video causally: users can start from an initial shot, issue new prompts during rollout, introduce new events or viewpoints, and continue generation without recomputing previous shots. Importantly, this capability is demonstrated at practical model scale. We build CausalCine on a 14B-parameter video generator and run it with streaming KV caching on 8 NVIDIA H200 GPUs at 16 FPS. This makes interactive multi-shot generation possible in real time, while preserving long-range semantic memory across shot boundaries. Experiments show that CausalCine substantially outperforms autoregressive baselines in shot-level quality, prompt alignment, identity preservation, and transition structure, and approaches the visual quality of bidirectional models while retaining the efficiency and interactivity unique to causal generation.

\section{Related Works}

\subsection{Autoregressive Video Generation}
Autoregressive video generation factorizes a video into sequentially generated frames or chunks, making it naturally suited for long-horizon rollout, KV-cache reuse, and interactive continuation~\cite{bruce2024genie,huang2025liveavatar,ki2026avatarforcing,shin2025motionstream,liu2026realwonder}. Recent autoregressive video models often start from pretrained diffusion models, then make them causal so that videos can be generated chunk by chunk~\cite{yin2025causvid,huang2025sf,zhu2026cf,yang2025longlive,liu2025rolling,cui2025sf++}. CausVid~\cite{yin2025causvid} distills bidirectional diffusion into a few-step causal model for low-latency streaming, while Self Forcing~\cite{huang2025sf} and Causal Forcing~\cite{zhu2026cf} reduce train--test mismatch by supervising the model on its own rollout distribution with distribution matching distillation~\cite{yin2024dmd,yin2024dmd2}. Long-context AR systems further extend generation through rolling caches, local windows, fixed anchors, or runtime prompt updates~\cite{yang2025longlive,liu2025rolling,yesiltepe2025infinity,cui2026lol}. However, these methods are primarily designed for single-scene continuation, where long video generation is treated as extending a local motion pattern. We study autoregressive generation in the multi-shot setting, where the model causally introduces new shots, prompts, and events while preserving long-range story context.

\subsection{Multi-Shot Video Generation}
Multi-shot video generation aims to synthesize coherent long videos with multiple shots, scene transitions, and evolving story structure. Existing approaches often decompose the task into scripts, shots, or keyframes, and generate each segment with a short-video model~\cite{bansal2024talc,hu2024storyagent,long2024videostudio,wang2025autostory,xie2024dreamfactory,zhao2024moviedreamer,xiao2025captain,zheng2024videogen,zhou2024storydiffusion}. These methods provide explicit control over story planning, but cross-shot consistency must be recovered through separate linking or refinement stages. More recent holistic methods model multiple shots jointly inside a unified diffusion process~\cite{meng2025holocine,cai2025mixture,jia2025moga,qi2025mask,wang2025echoshot,guo2025long}, improving global consistency by allowing all shots to interact during generation. However, their bidirectional formulation requires joint generation over all shots, leading to quadratic cost with video length and limiting online interaction. In contrast, CausalCine generates multi-shot videos autoregressively, allowing new prompts to be appended on the fly without recomputing previous content.

\subsection{Memory in Video Generation Models}
Memory mechanisms are widely used to extend video generation beyond the local temporal window. Streaming AR models typically retain recent frames together with fixed anchors or sink tokens from the sequence beginning~\cite{xiao2023efficient,yang2025longlive}, while other methods compress history into compact representations or maintain multi-scale short- and long-term memory~\cite{zhang2025frame,gu2025long,henschel2025streamingt2v,hong2024slowfast}. More recent work explores adaptive memory, retrieving history based on camera pose, field-of-view overlap, 3D scene structure, or content relevance~\cite{xiao2025worldmem,yu2025context,li2025vmem,cai2025mixture,ji2025memflow,guo2025end}. Inspired by these directions, we integrate content-aware memory retrieval directly into the visual KV cache, and show that such adaptive memory is effective for the more challenging setting of few-step causal multi-shot generation.
\section{Method}
\label{sec:method}

\begin{figure}[t]
    \centering
    \includegraphics[width=\linewidth]{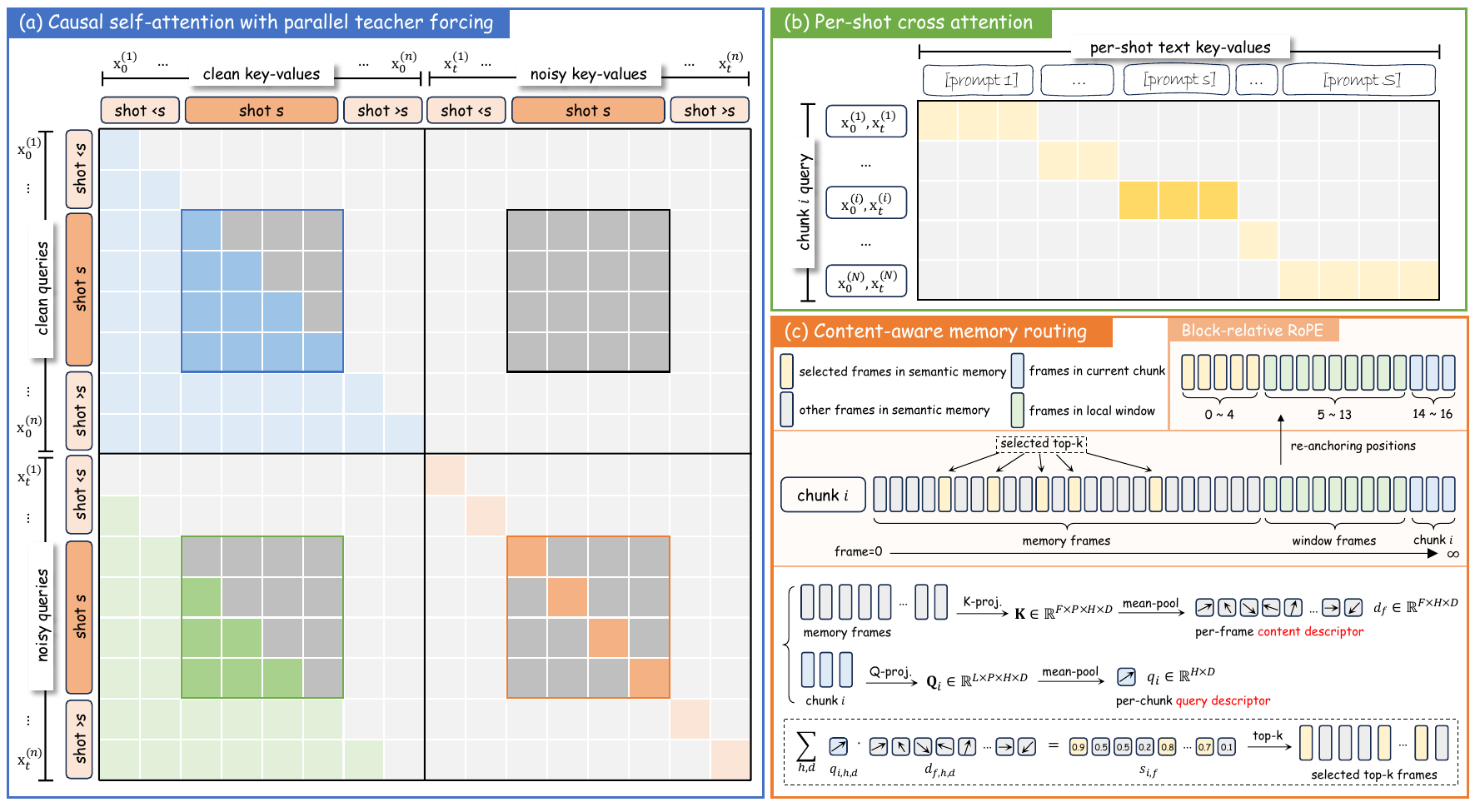}
    \caption{Overview of CausalCine. 
(a) A 2N-segment teacher-forcing layout trains causal multi-shot dependencies in one parallel forward pass. 
(b) Per-shot cross-attention routes each chunk to its active shot prompt. 
(c) Content-Aware Memory Routing retrieves relevant historical KV entries and applies Block-Relative RoPE to keep positional phases within the training range during long rollouts.}
\vspace{-1.5em}
    \label{fig:overview}
\end{figure}

We organize our framework around the design rationale that \emph{causality and multi-shot structure should be learned before step compression}. Starting from a pretrained bidirectional flow-matching video diffusion model, we (i) tune it into a full-step causal multi-shot generator with parallel teacher forcing on long cinematic videos (\cref{sec:method:tune}); (ii) replace its temporal positional heuristics with a content-aware memory router shared by training and inference (\cref{sec:method:memory}); and (iii) distill the resulting full-step causal model into a four-step generator for interactive synthesis (\cref{sec:method:dmd}).

\subsection{Preliminaries}
\label{sec:method:prelim}

\paragraph{Flow-matching video diffusion.}
We operate in the video VAE latent space, where a clean video latent $\mathbf{x}_0\!\in\!\mathbb{R}^{F\times C\times H\times W}$ and Gaussian noise $\boldsymbol{\epsilon}\!\sim\!\mathcal{N}(\mathbf{0},\mathbf{I})$ are interpolated as $\mathbf{x}_t=(1{-}\sigma_t)\mathbf{x}_0+\sigma_t\boldsymbol{\epsilon}$ under a shifted schedule~\cite{esser2024scaling}. A DiT \cite{peebles2023dit} velocity field $v_\theta(\mathbf{x}_t,t,\mathbf{c})$ is trained with the rectified flow-matching loss
\begin{equation}
    \mathcal{L}_{\text{FM}}
    = \mathbb{E}_{t,\mathbf{x}_0,\boldsymbol{\epsilon}}
      \big\Vert v_\theta(\mathbf{x}_t,t,\mathbf{c})
              - (\boldsymbol{\epsilon}-\mathbf{x}_0)\big\Vert^2,
\end{equation}
and sampling integrates $\mathrm{d}\mathbf{x}/\mathrm{d}t = v_\theta$ with a few-step Euler solver.

\paragraph{Distribution matching distillation.}
DMD~\cite{yin2024dmd,yin2024dmd2} compresses a pretrained teacher into a few-step student $G_\phi$ by minimizing a reverse KL between the student and teacher distributions at every noise level $t$, yielding the implicit gradient
\begin{equation}
    \nabla_\phi\mathcal{L}_{\text{DMD}}
    = \mathbb{E}_t\big[(s_{\text{fake}}(\mathbf{x}_t,t)
      - s_{\text{real}}(\mathbf{x}_t,t))\,
      \partial_\phi G_\phi\big],
      \label{eq:dmd_grad}
\end{equation}
where $s_{\text{real}}$  is predicted by the frozen teacher and $s_{\text{fake}}$ is predicted by an auxiliary score network co-trained with flow matching on the student's rollouts. We use this formulation in ~\cref{sec:method:dmd} and augment it with adversarial regularization.

\subsection{Long Multi-Shot Causal Tuning}
\label{sec:method:tune}

This stage converts a pretrained bidirectional video diffusion transformer into a causal generator whose training already covers the distribution of cinematic shot transitions, through a unified teacher-forcing regime on long multi-shot videos.

\paragraph{Causal chunk-wise formulation.}
We factorize a long video latent along the temporal axis into $N$ contiguous \emph{chunks} $\{\mathbf{x}^{(1)},\ldots,\mathbf{x}^{(N)}\}$ with $\mathbf{x}^{(i)}\!\in\!\mathbb{R}^{L\times C\times H\times W}$. A chunk is the \emph{unit of autoregression}, not a narrative unit; in our experiments $L{=}3$ latent frames ($\approx\!12$ video frames), and frame-wise AR is the special case $L{=}1$. The joint distribution factorizes causally,
\begin{equation}
    p_\theta(\mathbf{x}^{(1{:}N)} \mid \mathbf{c}_{1:N})
    = \prod_{i=1}^{N}
      p_\theta(\mathbf{x}^{(i)} \mid \mathbf{x}^{(<i)}, \mathbf{c}_i).
    \label{eq:causal}
\end{equation}
A multi-shot video consists of $S$ contiguous shots with prompts $\{\mathbf{c}_{(s)}\}_{s=1}^{S}$ separated by latent-frame boundaries $\mathcal{B}=\{b_1,\ldots,b_{S-1}\}$. The text condition for chunk $i$ is therefore \emph{shot-indexed}, $\mathbf{c}_i=\mathbf{c}_{(\pi(i))}$, where $\pi(i)\!\in\!\{1,\ldots,S\}$ is the shot containing chunk $i$. At a shot boundary the prompt $\mathbf{c}_i$ changes; the generated chunk $\mathbf{x}^{(i)}$ is then expected to faithfully reflect the new prompt rather than extrapolate the previous shot, a regime in which short clip-trained AR models tend to collapse onto static or looping content~\cite{cai2026mmm}.


\paragraph{Parallel teacher forcing with $2N$-segment packing.}
A step-by-step rollout of Eq.~\eqref{eq:causal} during training is prohibitive in time and memory. Following teacher-forcing training~\cite{huang2025sf,zhu2026cf}, we pack, for each video, a single $2N$-segment input of clean and noisy copies of all chunks:
\begin{equation}
    \mathbf{X}_{\text{TF}}
    = \big[
        \underbrace{\mathbf{x}^{(1)}_0,\ldots,\mathbf{x}^{(N)}_0}_{\text{clean context}},\;
        \underbrace{\mathbf{x}^{(1)}_t,\ldots,\mathbf{x}^{(N)}_t}_{\text{noisy queries}}
      \big].
\end{equation}
Clean segments carry timestep $0$; all noisy segments share a single sampled $t\!\sim\!p(\sigma)$, keeping the loss aligned across chunks. The block-sparse self-attention mask (\cref{fig:overview}(a)) has four quadrants: (a) clean$\to$clean is causal, where each clean chunk attends to itself and all preceding clean chunks; (b) noisy$\to$clean allows each noisy chunk to attend only to preceding clean chunks; (c) noisy$\to$noisy is restricted to the diagonal, ruling out leakage from future noisy chunks; and (d) clean$\to$noisy is fully masked. The flow-matching loss is computed on the noisy half:
\begin{equation}
    \mathcal{L}_{\text{tune}}
    = \mathbb{E}_{t,\mathbf{X}_{\text{TF}}}\,\frac{1}{N}\sum_{i=1}^{N}
      \big\Vert v_\theta(\mathbf{X}_{\text{TF}}; t, \mathcal{M})_{[N+i]}
      - (\boldsymbol{\epsilon}^{(i)}-\mathbf{x}^{(i)}_0)\big\Vert^2.
\end{equation}
This layout exposes the causal visibility pattern that the model uses with a KV cache at inference, while replacing sequential rollout with a single parallel forward pass. In practice, the noisy$\to$clean quadrant is further sparsified into a local window plus content-routed long memory, as in~\cref{sec:method:memory}.
\paragraph{Per-shot text conditioning.} As shown in~\cref{fig:overview}(a), given shot boundaries $\mathcal{B}$, both segments of chunk $i$ in the packed layout (the clean context $\mathbf{x}^{(i)}_0$ and the noisy query $\mathbf{x}^{(i)}_t$) are conditioned on the same shot prompt $\mathbf{c}_{(\pi(i))}$ via segment-level cross-attention; cross-attention between segments is forbidden, so each chunk only sees its own shot's prompt tokens. This explicit shot-indexed routing ties shot-boundary prompt changes to visual transitions.

\paragraph{Scaling to long cinematic videos.}
Short clips rarely span shot boundaries, failing to supervise transition dynamics or long range entity correlation, the very essence of cinematic videos. To learn these behaviors, we train natively on long multi-shot sequences of $\sim$15\,s ($\approx\!241$ video frames). This long-form supervision provides the critical signals needed for the causal model to actively introduce new scenes and preserve identities across cuts, rather than merely extrapolating a single shot. To make this extensive context tractable, our $2N$-packing trains all targets in a single parallel pass, while FSDP~\cite{zhao2023fsdp} and sequence-parallel attention~\cite{jacobs2023ulysses} absorb the $\mathcal{O}(NL)$ memory footprint.

\subsection{Content-Aware Memory Routing}
\label{sec:method:memory}

Long rollouts require compressing the growing KV cache into a bounded attention buffer. Prior AR video generators typically use position-defined memory, such as a local window plus first-frame sink tokens~\cite{xiao2023efficient,zhu2026cf}, which is fragile when multi-shot generation introduces new or reappearing content far from the opening frame. We instead augment the local window with content-addressable memory: each attention layer retrieves history frames whose keys best match the current query. The same routing module is used in TF training and AR inference, as shown in~\cref{fig:overview} (c).

\paragraph{Frame-level, chunk-shared routing.}
Let $\mathbf{K}\!\in\!\mathbb{R}^{F\!\times\!P\!\times\!H\!\times\!D}$ stack the cached keys of $F$ history latent frames, where $P$ is the number of spatial tokens per frame, $H$ the number of heads, and $D$ the head dimension. Following prior token-level sparse routing in language and video models~\cite{lu2025moba,cai2025mixture}, where mean-pooled keys have been shown to provide effective retrieval signals, for every cached frame $f$ we store a compact \emph{content descriptor} obtained by mean-pooling its key over spatial tokens,
\begin{equation}
    \mathbf{d}_f \,=\, \frac{1}{P}\sum_{p=1}^{P} \mathbf{K}_{f,p,:,:}
    \;\in\;\mathbb{R}^{H\times D}.
\end{equation}
For the current chunk $\mathbf{x}^{(i)}$ we form a query descriptor $\mathbf{q}_i\!\in\!\mathbb{R}^{H\times D}$ in the same way, mean-pooling the chunk's queries over both its $L$ frames and $P$ spatial tokens, so that all $L$ frames in chunk $i$ share one routing decision. We score every out-of-window history frame by a head-aggregated dot product,
\begin{equation}
    s_{i,f} \,=\, \sum_{h,d}\,
      \mathbf{q}_{i,h,d}\,\mathbf{d}_{f,h,d},
    \label{eq:desc_score}
\end{equation}
and select the top-$k$ frames. Letting $\mathcal{W}_i$ denote the local window of $W$ chunks preceding $i$ and $\mathcal{H}_i$ the out-of-window history, the effective receptive field of chunk $i$ is
\begin{equation}
    \mathcal{R}_i
    \,=\, \underbrace{\mathrm{Top}\text{-}k\bigl(\{s_{i,f}\}_{f\in\mathcal{H}_i}\bigr)}_{\text{semantic memory}}
    \;\cup\;    
    \underbrace{\mathcal{W}_i}_{\text{local window}}
    \;\cup\;\{\mathrm{\text{current chunk}}\}.
    \label{eq:route_field}
\end{equation}
We use $W{=}3$ chunks and $k{=}5$ frames throughout. The routing is model-adaptive but parameter-free. Although the top-$k$ selection is not differentiable, scores are computed from the learned query/key representations. Routing is applied to self-attention only; cross-attention to text remains as in ~\cref{sec:method:tune}.

\paragraph{Block-Relative RoPE.}
Content-based routing may retrieve frames beyond the training horizon $F_\mathrm{train}$, e.g.,1000th frame in a minute-long rollout. Applying 3D RoPE at these global positions exposes attention to unseen phases and may cause severe visual artifacts.

We avoid this by re-anchoring positions after retrieval. Keys are stored unrotated in the cache; after top-$k$ selection, RoPE is applied to the selected memory, local window, and current chunk using compact block-relative positions:
\begin{equation}
    \underbrace{[\,0,\ldots,k{-}1\,]}_{\text{memory}}
    \Vert
    \underbrace{[\,k,\ldots,k{+}WL{-}1\,]}_{\text{window}}
    \Vert
    \underbrace{[\,k{+}WL,\ldots,k{+}(W{+}1)L{-}1\,]}_{\text{current}},
\end{equation}
whose span is $k+(W{+}1)L \le F_\mathrm{train}$ by construction ($5+4{\cdot}3=17 \ll 61$ in our setting). Since the same cached key may receive different relative positions for different queries, keys cannot be rotated once at write time. This Block-Relative RoPE keeps all attention phases within the training-range envelope regardless of rollout length.

\subsection{Few-Step Causal Distillation}
\label{sec:method:dmd}

With the causal multi-shot model from~\cref{sec:method:tune} and the memory router from~\cref{sec:method:memory}, we distill the many-step flow-matching teacher into a four-step autoregressive generator $G_\phi$ using Distribution Matching Distillation (DMD)~\cite{yin2024dmd,yin2024dmd2} and an adversarial objective. The distilled student preserves the causal chunk-wise architecture, per-shot conditioning, without modification.

\paragraph{Teacher Forcing causal ODE initialization.}
Before self-forced DMD, we initialize the student via causal ODE distillation~\cite{zhu2026cf}. Given a ground-truth history $\mathbf{x}^{(<i)}_{\mathrm{gt}}$ and shot prompt $\mathbf{c}_i$, we generate a teacher PF-ODE trajectory $\mathbf{z}^{(i)}_{\tau}$ from noise $\boldsymbol{\epsilon}^{(i)}$ (subsampled to 4 steps $\tau\!\in\!\mathcal{S}$ from a 48-step solver). We train the student to predict the teacher's final denoised output $\mathbf{z}^{(i)}_0$ by minimizing:
\begin{equation}
    \mathcal{L}_{\mathrm{init}}
    =
    \mathbb{E}_{i,\tau\sim\mathcal{S}}
    \left\|
        \hat{\mathbf{x}}_{0,\phi}
        \left(
            \mathbf{z}^{(i)}_{\tau},
            \mathbf{x}^{(<i)}_{\mathrm{gt}},
            \tau,
            \mathbf{c}_i
        \right)
        - \mathbf{z}^{(i)}_0
    \right\|_2^2.
\end{equation}
This aligns the few-step student with the teacher's causal visibility pattern, which is crucial for preventing unstable targets during the subsequent self-forced training where the teacher's scores are queried on the student's own long-horizon rollouts.

\paragraph{Distribution matching distillation with adversarial regularization.}
We further refine $G_\phi$ under a self-forcing framework~\cite{huang2025sf}: each update starts from the student's own causal rollout $\tilde{\mathbf{x}}_{0,\phi}$ using the inference KV cache and memory routing. After perturbing it to $\tilde{\mathbf{x}}_{t,\phi}$, we apply the DMD gradient (\cref{eq:dmd_grad}). The frozen real denoiser $T_\psi$ and the flow-matching-updated auxiliary fake denoiser $T_{\phi^-}$ are initialized from our tuned multi-shot model. 

To reduce sequence-level drift in long rollouts, we follow APT~\cite{lin2025apt} and attach a lightweight GAN head $D_\eta$ to the intermediate features of $T_{\phi^-}$. Let $d_\eta(\mathbf{y}_t) = D_\eta(F_{\phi^-}(\mathbf{y}_t,t,\mathbf{c}))$ denote the logit. We optimize the standard logistic adversarial loss:
\begin{equation}
    \mathcal{L}_{D} = \mathbb{E}_{\mathbf{x}_0}[f(-d_\eta(\mathbf{x}_{t}))] + \mathbb{E}_{\tilde{\mathbf{x}}_{0,\phi}}[f(d_\eta(\tilde{\mathbf{x}}_{t,\phi}))],
    \quad
    \mathcal{L}_{G} = \mathcal{L}_{\mathrm{DMD}} + \lambda_{\mathrm{adv}}\mathbb{E}_{\tilde{\mathbf{x}}_{0,\phi}}[f(-d_\eta(\tilde{\mathbf{x}}_{t,\phi}))],
\end{equation}
where $f(u)=\log(1+\exp(u))$. The generator is trained with $\mathcal{L}_{G}$ and the discriminator with $\mathcal{L}_{D}$, effectively penalizing drift in camera motion and subject framing.

\section{Experiments}
\label{sec:exp}

\noindent{\textbf{Implementation Details.}}
We build our autoregressive framework on Wan2.1-T2V-14B~\cite{wan2025wan} to generate videos at resolution $832\times480$.  The causal base model is trained with chunk-wise teacher forcing on 100k long multi-shot videos, where each chunk contains three latent frames. The training process is conducted on 64 NVIDIA H800 GPUs. At inference time, the model generates chunks sequentially with KV caching. Our distilled student uses four denoising steps and inherits the same per-shot text routing and memory mechanism as the causal base.

\noindent{\textbf{Evaluation Protocol.}} Following \citet{meng2025holocine}, we use Gemini 2.5 Pro~\cite{team2023gemini,comanici2025gemini} to build a $100$-prompt multi-shot benchmark. Each prompt contains a global story description, five shot-level captions, and target shot-cut locations, covering character reappearance, scene changes, shot-reverse-shot interactions, viewpoint changes, and long temporal gaps. Following VBench~\cite{huang2024vbench}, we evaluate visual quality, prompt following, temporal consistency, long-range consistency, and shot structure. Specifically, we report LAION aesthetic score~\cite{schuhmann2022laion}, shot-level ViCLIP text-video similarity~\cite{wang2023internvid,wang2022internvideo}, within-shot subject/background consistency using DINO~\cite{caron2021emerging} and CLIP~\cite{radford2021learning}, inter-shot character consistency using DINOv2~\cite{oquab2023dinov2} on matched pairs, and shot-cut accuracy (SCA)~\cite{meng2025holocine} by matching TransNetV2~\cite{soucek2024transnet}-detected cuts to target boundaries. To ensure a fair comparison, all baselines generate videos under identical settings as ours, using the same set of prompts, resolution, and length.

\begin{figure}[tbp]
    \centering
    \vspace{-0.5em}
    \includegraphics[width=0.95\linewidth]{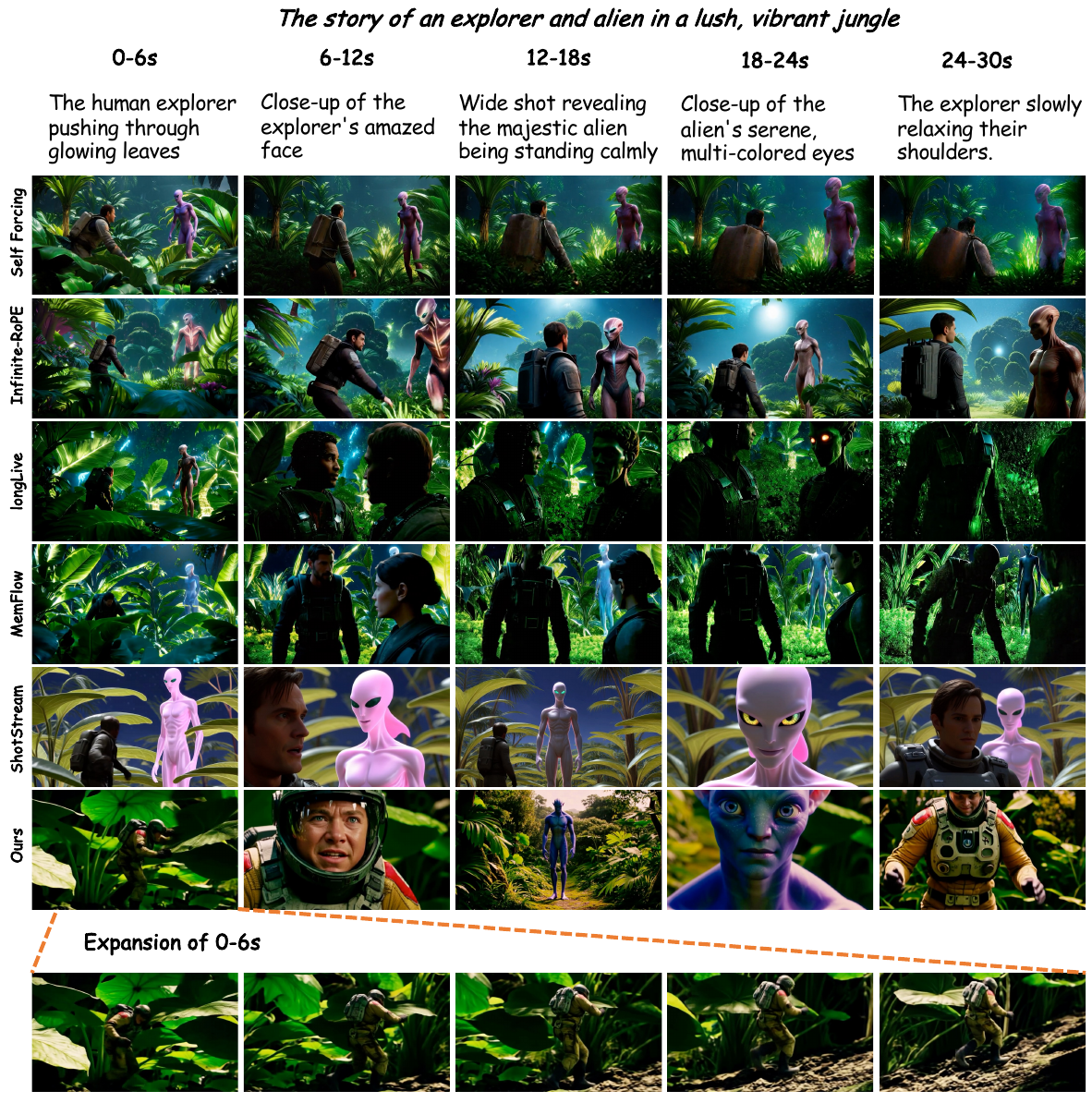}
    \vspace{-0.8em}
    \caption{Comparison with autoregressive and streaming long-video baselines. Existing autoregressive methods often remain tied to the initial scene, repeat similar compositions, or miss requested viewpoint changes. CausalCine better follows the shot progression while preserving subjects across shots; the expanded first shot shows coherent intra-shot motion.}
    \label{fig:comparison}
    \vspace{-1em}
\end{figure}

\begin{figure}[!htbp]
    \centering
    \includegraphics[width=0.95\linewidth]{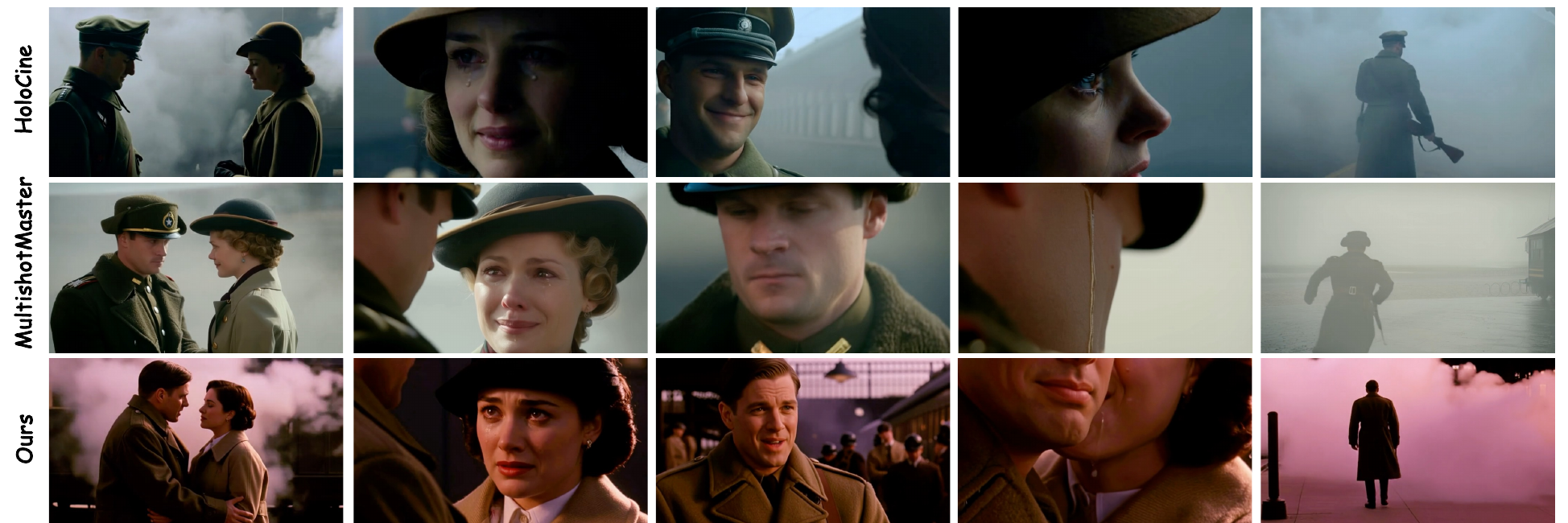}
    \caption{Our causal generator produces results comparable to bidirectional baselines.}
    \label{fig:compare_bid}
\end{figure}

\paragraph{Comparisons.} We first compare with autoregressive long-video generation methods, including Self-Forcing~\cite{huang2025sf}, Infinity-RoPE~\cite{yesiltepe2025infinity}, LongLive~\cite{yang2025longlive}, MemFlow~\cite{ji2025memflow}, and ShotStream~\cite{luo2026shotstream}. These methods extend generation through causal rollout, KV caching, or long-context positional extrapolation, but most of them are primarily designed for short-context continuation. As shown in~\cref{tab:main_ar,fig:comparison}, they often produce locally smooth videos that remain semantically static, repeating similar layouts or missing requested shot-level changes. Our method achieves the best overall performance, with clear gains in text alignment and shot-cut accuracy, showing stronger ability to follow changing per-shot instructions while preserving subject consistency.

We also compare with bidirectional multi-shot models~\cite{meng2025holocine,wang2025multishotmaster}, which generate the full sequence jointly. Note that, to align with the preferred generation length of these bidirectional baselines, we evaluate this comparison under their 15s setting. As shown in~\cref{tab:bidir,fig:compare_bid}, our causal generator achieves comparable visual quality and cross-shot coherence, while being substantially faster at inference. In addition, our method naturally supports interactive continuation, where users can append new shot prompts during generation without providing the entire prompt sequence in advance.

\begin{table}[tbp]
\centering
\small %
\caption{Comparison with autoregressive video generation baselines. 
Best values per column are in \textbf{bold} and second best are \underline{underlined}. Our method achieves the best overall performance.}
\label{tab:main_ar}
\setlength{\tabcolsep}{4pt} 
\begin{tabular}{lccccc}
\toprule

Method & Aesthetic $\uparrow$ & \makecell{Text \\ Align. $\uparrow$} & \makecell{Subject \\ Consistency $\uparrow$} & \makecell{Background \\ Consistency $\uparrow$} & \makecell{SCA. $\uparrow$} \\
\midrule
Self-Forcing      & \underline{0.6228} & 0.1395          & \underline{0.9668} & \textbf{0.9717} & 0.5052 \\
Infinity-RoPE     & 0.6225          & 0.1716          & 0.8609          & 0.9091          & 0.7842 \\
LongLive          & 0.6198          & 0.1552          & 0.9319          & 0.9487          & 0.5021 \\
MemFlow           & 0.6139          & 0.1587          & 0.9293          & 0.9483          & 0.5092 \\
ShotStream        & 0.6146          & \underline{0.1753} & 0.9617          & 0.9670          & \underline{0.9647} \\
\midrule
Ours & \textbf{0.6261} & \textbf{0.1980} & \textbf{0.9717} & \underline{0.9675} & \textbf{0.9732} \\
\bottomrule
\end{tabular}
\vspace{-1em}
\end{table}
\begin{table}[tbp]

\centering
\small

\caption{Comparison with bidirectional multi-shot generation models under the 15-second setting.}
\label{tab:bidir}
\setlength{\tabcolsep}{4pt}
\begin{tabular}{lcccccccc}
\toprule
\multirow{2}{*}{Method} & \multirow{2}{*}{Architecture} & \multirow{2}{*}{Aesthetic $\uparrow$} & \multirow{2}{*}{\makecell{Text \\ Align. $\uparrow$}} & \multicolumn{2}{c}{Intra-Shot Cons. $\uparrow$} & \multirow{2}{*}{\makecell{Inter-Shot \\ Cons. $\uparrow$}} & \multirow{2}{*}{\makecell{SCA $\uparrow$}} \\
\cmidrule(lr){5-6}
& & & & Subject & Background & & \\
\midrule
HoloCine & Bidirectional & 0.5842 & \textbf{0.2050} & 0.9728 & 0.9711 & \textbf{0.6821} & 0.9694 \\
MultiShotMaster & Bidirectional & 0.5811 & 0.2046 & 0.9626 & 0.9671 & 0.6530 & 0.9678 \\
\midrule
Ours & Causal, 4step & \textbf{0.6194} & 0.2004 & \textbf{0.9823} & \textbf{0.9752} & 0.6608 & \textbf{0.9883} \\
\bottomrule
\end{tabular}
\end{table}

\subsection{Ablation on Key Design Choices}
\begin{figure}[h]
    \centering
    \vspace{-4pt}
    \includegraphics[width=0.95\linewidth]{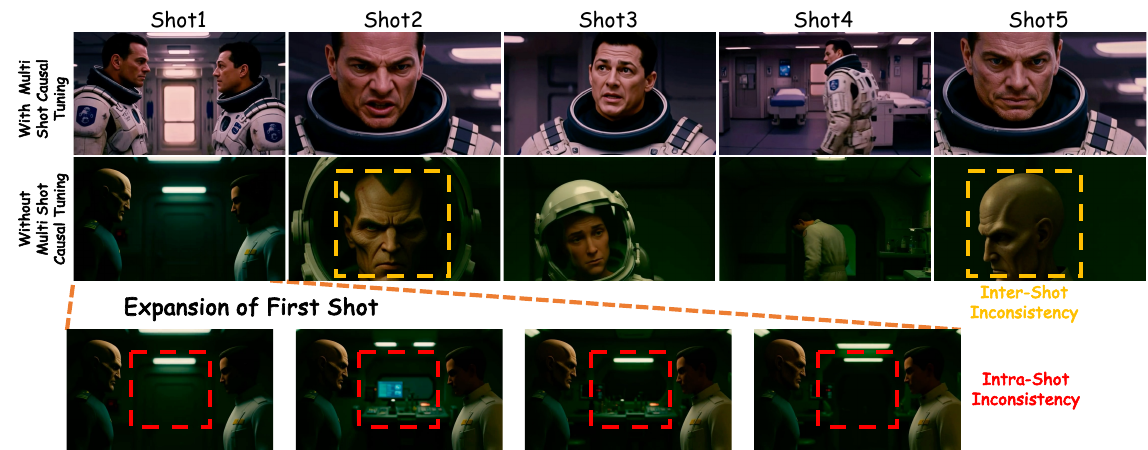}
\caption{Effect of learning causal multi-shot structure before step compression. Directly initializing a few-step student from a short bidirectional teacher leads to unstable intra-shot content and inconsistent identities across shots. Our full pipeline first learns long-context causal multi-shot generation and then compresses sampling steps, improving temporal stability and cross-shot identity preservation.}
    \label{fig:long_tuning}
\end{figure}

\textbf{Ablation on Long Multi-Shot Causal Tuning}
\label{sec:exp:long_tuning}
\begin{table}[tbp]
\centering
\small
\caption{Ablation studies on causal tuning and memory design. }
\label{tab:ablations}
\setlength{\tabcolsep}{4pt}
\begin{tabular}{llcccccc}
\toprule
\multirow{2}{*}{Ablation} & \multirow{2}{*}{Method}
& \multirow{2}{*}{Aesthetic $\uparrow$}
& \multirow{2}{*}{\makecell{Text \\ Align. $\uparrow$}}
& \multicolumn{2}{c}{Intra-Shot Cons. $\uparrow$}
& \multirow{2}{*}{\makecell{Inter-Shot \\ Cons. $\uparrow$}}
& \multirow{2}{*}{SCA $\uparrow$} \\
\cmidrule(lr){5-6}
& & & & Subject & Background & & \\
\midrule
\multirow{2}{*}{Causal tuning}
& w/o multi-shot tuning & 0.5967 & 0.1921 & 0.9311 & 0.9519 & 0.5034 & 0.5042 \\
& w/ multi-shot tuning  & \textbf{0.6261} & \textbf{0.1980} & \textbf{0.9717} & \textbf{0.9675} & \textbf{0.6529} & \textbf{0.9732} \\
\midrule
\multirow{3}{*}{Memory}
& w/o memory & 0.5827 & 0.2181 & 0.9432 & 0.9412 & 0.5832 & \textbf{0.9772} \\
& first-frame sink & \textbf{0.6017} & 0.2285 & 0.9575 & 0.9443 & 0.6106 & 0.9618 \\
& content routing (ours) & 0.5974 & \textbf{0.2394} & \textbf{0.9628} & \textbf{0.9529} & \textbf{0.7530} & 0.9745 \\
\bottomrule
\end{tabular}
\end{table}
We ablate the ordering of causal multi-shot learning and step compression. Our full framework first adapts the bidirectional video model into a long-context causal multi-shot model, and then performs ODE initialization and DMD distillation for few-step generation. In the ablated setting, we skip this long multi-shot causal tuning stage and directly perform ODE initialization by aligning the student to trajectories from the original 5s bidirectional model. The following DMD stage, student architecture, and inference procedure are kept the same.
\Cref{tab:ablations} shows that direct compression from the 5s bidirectional model degrades prompt following, shot-cut control, and long-range consistency. This indicates that step compression cannot reliably recover causal multi-shot behavior when it is absent from the initialization. As shown in~\cref{fig:long_tuning}, the ablated model suffers from unstable intra-shot content and inconsistent cross-shot identity, while our full pipeline remains more stable.

\begin{figure}[htbp]
    \centering
    \vspace{-5pt}
    \includegraphics[width=0.95\linewidth]{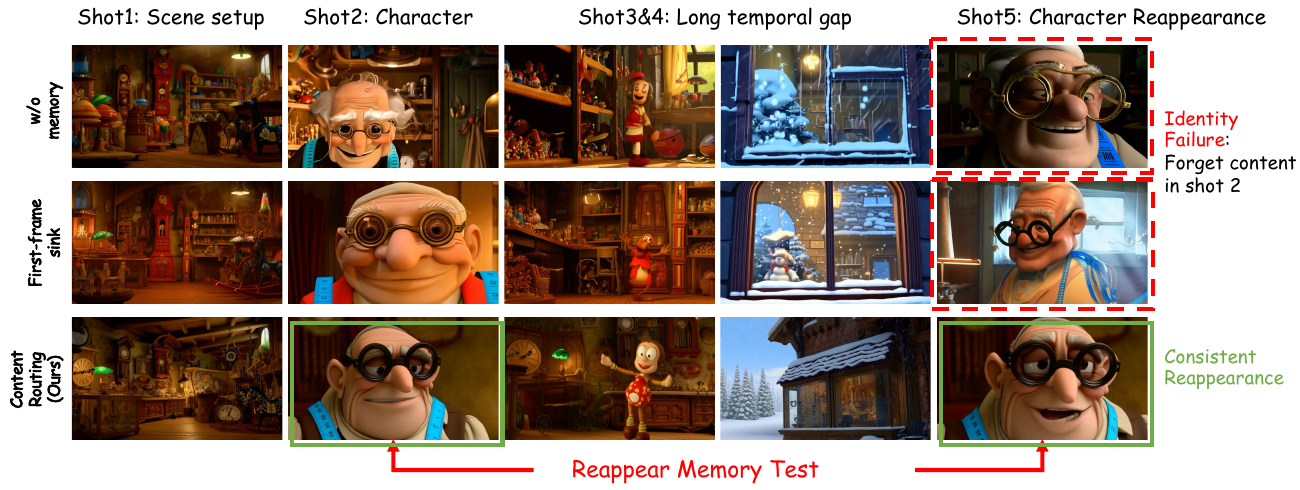}
    \caption{Content-aware memory routing better preserves character identity across long temporal gaps than no-memory and first-frame sink variants, enabling a consistent reappearance in final shot.}
    \label{fig:ablation_memory}
    \vspace{-8pt}
\end{figure}

\textbf{Ablation on Memory Design}
\label{sec:exp:memory_ablation}
We study how memory design affects long multi-shot rollout. To better evaluate the memory mechanism, we construct a dedicated evaluation set comprising 100 memory test prompts generated by Gemini 2.5 pro~\cite{comanici2025gemini}, which specifically emphasize scenarios where subjects disappear and reappear across shots. \Cref{tab:ablations} compares three variants: removing long-term memory, using first-frame sink memory, and our content-aware memory routing. Without long-term memory, the model mainly relies on the local KV window and often forgets entities after long temporal gaps. First-frame sink memory provides a fixed positional anchor, but the earliest frames are not necessarily relevant after several shot cuts. Content-aware memory routing achieves the best inter-shot consistency by retrieving historical frames according to semantic affinity rather than temporal position. As shown in~\cref{fig:ablation_memory}, both the no-memory and first-frame-sink variants fail to faithfully recover the character when it reappears in the final shot. In contrast, our method retrieves the earlier character shot and preserves distinctive identity cues. 

\section{Conclusion}
We presented CausalCine, a causal framework for interactive multi-shot video generation. By learning long-form shot transitions before distillation and routing KV memory by content, CausalCine supports shot-level prompt updates, bounded-attention rollout, and cross-shot recall without regenerating earlier video. Experiments show improved prompt following, shot-cut control, and identity preservation over autoregressive baselines, with visual quality competitive with bidirectional multi-shot models.
\newpage
{
    \newpage
    \small
    \bibliographystyle{plainnat}
    \bibliography{main}
}
\newpage

\appendix

\setcounter{figure}{0}
\setcounter{table}{0}
\setcounter{equation}{0}

\renewcommand{\thefigure}{S\arabic{figure}}
\renewcommand{\thetable}{S\arabic{table}}
\renewcommand{\theequation}{S\arabic{equation}}

\renewcommand{\theHfigure}{supp.figure.\arabic{figure}}
\renewcommand{\theHtable}{supp.table.\arabic{table}}
\renewcommand{\theHequation}{supp.equation.\arabic{equation}}

\section{More Results}
\label{sec:more_results}

To provide a more complete view of the generated videos beyond the still frames shown in the main paper, we include additional \textbf{video results} in the supplementary material. These examples cover diverse multi-shot prompts, including character reappearance, viewpoint changes, scene transitions, and long-range cross-shot consistency.

We also include a recorded real-time interactive generation demo. The demo shows how CausalCine generates a video chunk by chunk, accepts newly appended shot-level prompts during rollout, and continues the sequence without regenerating previous shots.

For convenient browsing, the supplementary material contains an HTML gallery that organizes all video results and the interactive demo in one place. Readers can open the HTML file directly to view the results without clicking each video file individually.

\section{Causal Base Model vs. Four-Step Student}
\label{sec:exp:dmd}

\begin{figure}[htbp]
    \centering
    \includegraphics[width=\linewidth]{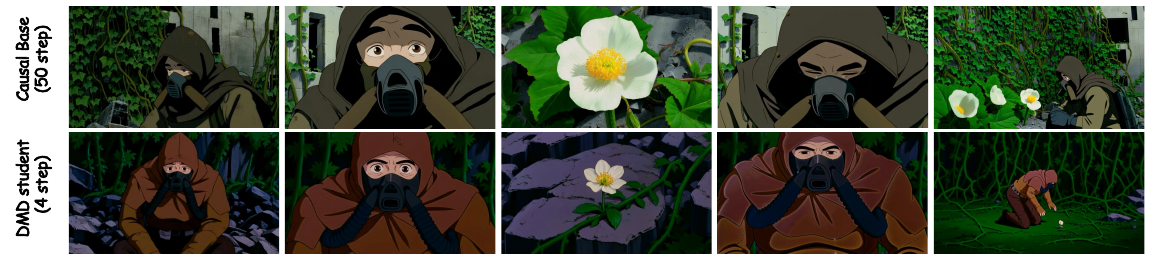}
    \caption{The four-step DMD student preserves the shot-level structure, subject identity, and visual composition of the 50-step causal base while substantially reducing the number of denoising steps.}
    \label{fig:ablation_50_step}
\end{figure}

We compare the full-step causal base model with the four-step DMD student to evaluate how much quality is retained after acceleration. The causal base is trained by long multi-shot teacher forcing and serves as the high-quality autoregressive teacher for distillation. As shown in~\cref{tab:base_vs_student,fig:ablation_50_step}, it already produces coherent multi-shot rollouts with clear shot transitions, strong prompt following, and stable recurring subjects.

The four-step student preserves these properties while reducing inference cost. Quantitatively, the student remains close to the causal base on visual quality, text alignment, consistency, and shot-cut accuracy. Qualitatively, ~\cref{fig:ablation_50_step} shows that the student follows the same shot progression and maintains the main subject across the sequence, despite using only four denoising steps. This indicates that DMD successfully compresses the denoising trajectory while maintaining the multi-shot structure, validating our design of first learning causal long-form behavior in the base model before distillation.

\begin{table}[htbp]
\centering
\small
\caption{Comparison between the causal base model and the four-step distilled student.}
\label{tab:base_vs_student}
\setlength{\tabcolsep}{4pt}
\begin{tabular}{lcccccccc}
\toprule
\multirow{2}{*}{Method} & \multirow{2}{*}{Steps} & \multirow{2}{*}{Aesthetic $\uparrow$} & \multirow{2}{*}{\makecell{Text \\ Align. $\uparrow$}} & \multicolumn{2}{c}{Intra-Shot Cons. $\uparrow$} & \multirow{2}{*}{\makecell{Inter-Shot \\ Cons. $\uparrow$}} & \multirow{2}{*}{\makecell{SCA $\uparrow$}} \\
\cmidrule(lr){5-6}
& & & & Subject & Background & & \\
\midrule
Causal base & 50 & 0.5930 & \textbf{0.2016} & 0.9628 & 0.9619 & \textbf{0.6621} & 0.9605 \\
DMD student & 4  & \textbf{0.6261} & 0.1980 & \textbf{0.9717} & \textbf{0.9675} & 0.6529 & \textbf{0.9732} \\
\bottomrule
\end{tabular}
\end{table}

\section{Effect of Adversarial Regularization}
\label{sec:appendix:gan_ablation}
\begin{figure}[htbp]
    \centering
      
    \includegraphics[width=\linewidth]{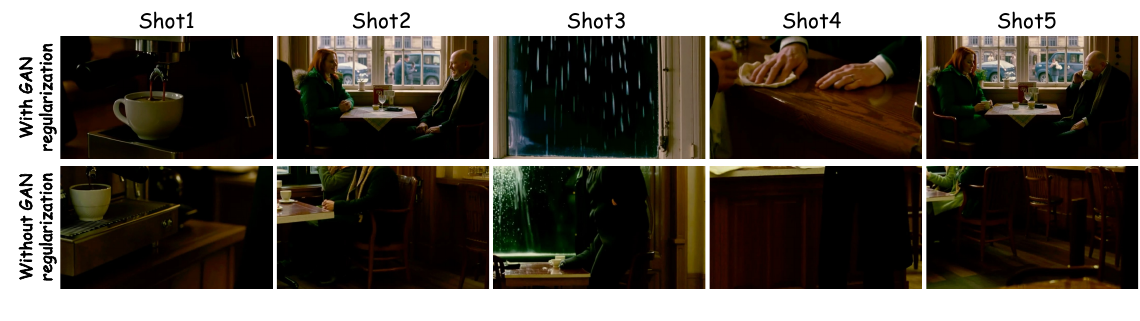}
    \caption{Ablation of adversarial regularization in DMD distillation. The student trained with the GAN regularization produces more stable subject framing and more plausible camera motion, while the model without GAN regularization shows sequence-level drift and irregular camera motion.}
  \label{fig:gan_ablation}
\end{figure}

We ablate the lightweight adversarial regularization used during DMD distillation by comparing the four-step student trained with and without the GAN head. As shown in~\cref{fig:gan_ablation}, the adversarial regularization stabilizes the sequence-level spatial distribution of long causal rollouts. 
Without the GAN head, the student still follows the multi-shot prompt, but with noticeable drift in camera motion and subject framing. In particular, recurring subjects may move away from the center of the frame, and some shots exhibit unnatural spatial shifts or composition changes. 
With GAN regularization, the generated shots maintain plausible camera motion and more stable subject placement.

\section{Limitations and Failure Case}
\label{sec:limitations}

\textbf{Limitations}. CausalCine targets a challenging setting: real-time autoregressive generation of coherent multi-shot videos with prompt changes, shot transitions, and long-range entity recall. To achieve multi-shot fidelity, we use Wan2.1-T2V-14B as the backbone rather than a smaller 1.3B-scale model commonly used by several prior autoregressive video systems, which also increases inference cost. With our distributed deployment, CausalCine reaches real-time generation at 16 FPS on 8 NVIDIA H200 GPUs, but this is still beyond the capability of consumer-grade GPUs. We view this mainly as a systems and model-scaling limitation rather than a fundamental limitation of the causal formulation: future smaller video backbones, quantization, faster attention kernels, and more optimized serving infrastructure could reduce the hardware requirement.



\begin{figure}[!ht]
    \centering
        
    \includegraphics[width=\linewidth]{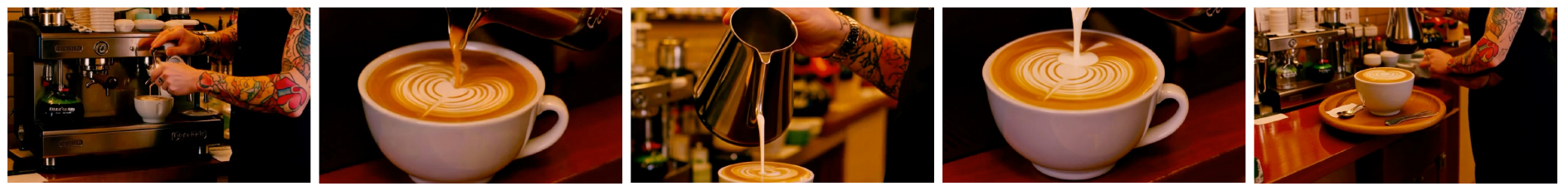}
    \caption{Failure case on fine-grained physical-state continuity. 
    CausalCine produces visually plausible coffee-making shots, but the milk stream, pitcher pose, hand position, and latte-art pattern do not evolve as a single physically consistent action across cuts.}
   \label{fig:failure_cases}

\end{figure}

\textbf{Failure case}. A typical failure case is fine-grained physical-state continuity across cuts. 
CausalCine is designed to preserve high-level narrative context, shot-level prompts, and recurring entities, but it does not explicitly maintain a structured state for small objects, contact geometry, or ongoing physical interactions. 
As a result, the model can generate individually plausible shots that do not compose into a fully consistent physical process. 
For example, in the coffee-making failure case in~\cref{fig:failure_cases}, the scene, cup, and latte-art theme remain recognizable, but the milk stream, pitcher position, hand pose, and foam pattern change in ways that are not physically continuous across cuts. 
This suggests that content-aware KV memory helps recall visual evidence, but does not by itself solve precise object-state tracking or action-level causality. 
Future work could combine causal video generation with explicit object-state memory, action constraints, or 3D-aware representations.


\end{document}